\documentclass[conference]{IEEEtran}
\usepackage[utf8]{inputenc}
\usepackage[numbers, square, sort&compress]{natbib} 

\usepackage{amsmath,amssymb,amsfonts}
\usepackage{graphicx}
\usepackage{textcomp}
\usepackage{xcolor}
\usepackage{setspace}
\usepackage{caption}
\usepackage{subcaption}
\usepackage{booktabs}
\usepackage{tabularx}
\usepackage{multicol}
\usepackage{tikz}
\usepackage{multirow}
\usepackage{algorithm, algpseudocode}
\usepackage{adjustbox}
\usepackage[font=small]{caption}
\newcolumntype{P}[1]{>{\centering\arraybackslash}p{#1}}
\newcolumntype{M}[1]{>{\centering\arraybackslash}m{#1}}
\usepackage{blindtext}
\usepackage{fancyhdr}
\usepackage{epstopdf}
 \usepackage{amsmath}
 \usepackage{authblk}
\IEEEoverridecommandlockouts

\begin{document}
\title{Advancing IIoT with Over-the-Air Federated Learning: The Role of Iterative Magnitude Pruning
\thanks{(\textit{Corresponding author: Aryan Kaushik})}
\thanks{This work was supported by the Natural Sciences and Engineering
Research Council of Canada (NSERC) under Grant RGPIN-2021-04050. The authors would also like to thank Denvr Dataworks, Calgary, Canada for their high-performance compute used to conduct this research.
}

\author{Fazal Muhammad Ali Khan, Hatem Abou-Zeid, Aryan Kaushik, Syed Ali Hassan
\thanks{Fazal Muhammad Ali Khan is with the School of Electrical Engineering \& Computer Science (SEECS), National University of Sciences \& Technology (NUST), Islamabad, Pakistan, and also a visiting research scholar at the Electrical and Software Engineering Department, University of Calgary, Calgary, Alberta, Canada (email: fkhan.phd20seecs@seecs.edu.pk).}
\thanks{Hatem Abou-Zeid is with the Electrical and Software Engineering Department, University of Calgary, Calgary, Alberta, Canada (email: hatem.abouzeid@ucalgary.ca).}
\thanks{Aryan Kaushik is with the School of Engineering and Informatics, University of Sussex, Brighton, UK, (email: aryan.kaushik@sussex.ac.uk).}
\thanks{Syed Ali Hassan is with the School of Electrical Engineering \& Computer Science (SEECS), National University of Sciences \& Technology (NUST), Islamabad, Pakistan (email: ali.hassan@seecs.edu.pk).}
}}

\maketitle

\begin{abstract}
The industrial Internet of Things (IIoT) under Industry 4.0 heralds an era of interconnected smart devices where data-driven insights and machine learning (ML) fuse to revolutionize manufacturing. A noteworthy development in IIoT is the integration of federated learning (FL), which addresses data privacy and security among devices. FL enables edge sensors, also known as peripheral intelligence units (PIUs) to learn and adapt using their data locally, without explicit sharing of confidential data, to facilitate a collaborative yet confidential learning process. However, the lower memory footprint and computational power of PIUs inherently require deep neural network (DNN) models that have a very compact size. Model compression techniques such as pruning can be used to reduce the size of DNN models by removing unnecessary connections that have little impact on the model's performance, thus making the models more suitable for the limited resources of PIUs. Targeting the notion of compact yet robust DNN models, we propose the integration of iterative magnitude pruning (IMP) of the  DNN model being trained in an over-the-air FL (OTA-FL) environment for IIoT. We provide a tutorial overview and also present a case study of the effectiveness of IMP in OTA-FL for an IIoT environment. Finally, we present future directions for enhancing and optimizing these deep compression techniques further, aiming to push the boundaries of IIoT capabilities in acquiring compact yet robust and high-performing DNN models.
\end{abstract}
 \begin{IEEEkeywords}
 Federated learning (FL), over-the-air (OTA) computation, DNNs, industrial IoT (IIOT), compression.
 \end{IEEEkeywords}

\section{Introduction}
The fourth phase of the industrial revolution, known as Industry 4.0, focuses on the digital overhaul across multiple fields, including smart manufacturing, advanced automation, and the aerospace industry. The aim is to integrate cutting-edge technologies to enhance efficiency and innovation, ushering a new level of connectivity and intelligence within the industries. This would foster the development of smart factories that can self-monitor, self-diagnose, and even predict future issues, leading to minimal downtime and increased productivity \cite{ziweiroadtoindustry4.0}. 

Central to this transformation are intelligent edge devices, such as sensors and actuators, also known as peripheral intelligence units (PIUs), which are deployed extensively to gather, process, and disseminate information. Therefore, the core of these advancements lies in the PIUs, which facilitate predictive maintenance, energy management, and automated quality control in the industry. 

The cornerstone of these devices becoming intelligent is the advent of machine learning (ML) models and deep neural networks (DNNs) that enable these devices to learn from past data, predict future conditions, and make informed decisions autonomously. As a result, PIUs have transitioned from being mere data collectors to smart, decision-making entities. This shift enhances the efficiency, reliability, and innovation of industrial operations, marking a significant milestone towards fully autonomous and intelligent industrial systems. Despite these benefits, the growth of PIUs and adoption of intelligence in the IIoT landscape face significant challenges due to data security, high-compute requirements of deep neural networks, and high energy consumption.

DNN model compression techniques can however be adopted to reduce the complexity and storage requirements of large DNNs used in IIoT systems. Model pruning is one of the primary methods to achieve this by identifying and eliminating unnecessary or less important connections within the DNN. The pruning process can be conducted iteratively and can lead to significant reductions in model size and computational complexity without impacting model accuracy. This enables the widespread deployment of advanced AI systems on PIUs that have limited resources.

\begin{figure*}[htbp]
\renewcommand{\figurename}{\hspace{0cm}Figure}
\centering
\begin{tikzpicture}
\node[draw, inner sep=3pt] {
    \includegraphics[width=\textwidth]{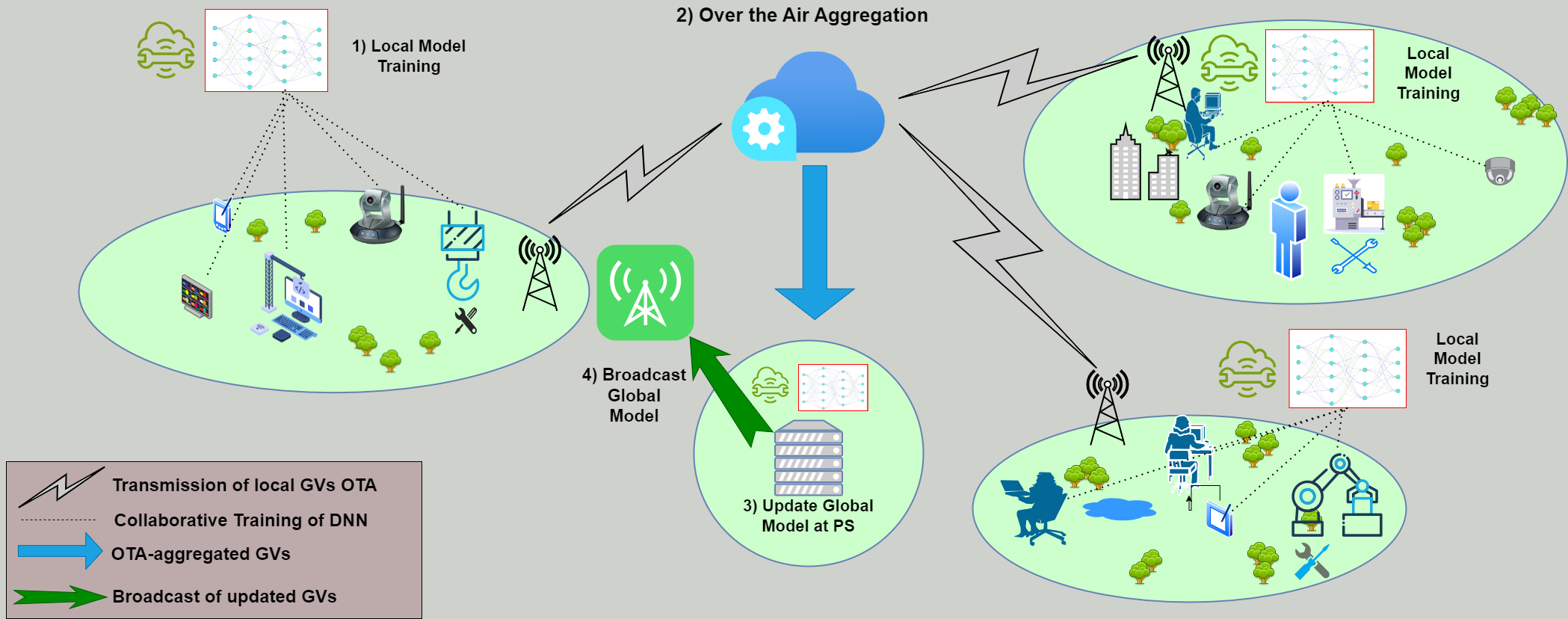}
};
\end{tikzpicture}
\caption{A depiction of OTA FL in IIoT. The model updates of PIU obtained via training with available datasets are transmitted and aggregated over a wireless medium, leveraging the superposition property of wireless channels. This simultaneous transmission and aggregation significantly reduces communication overhead and latency. The process ensures privacy and efficiency, making it ideal for real-world applications where data privacy is crucial and bandwidth is limited.}
\label{Figure main: Federated Learning via Over-the-air Aggregation}
\end{figure*}
Federated learning (FL) is a privacy-preserving distributed learning approach where model training occurs on decentralized devices. Each PIU trains a local model using its data, and only the \emph{model parameters} are transmitted to a central server for aggregation toward a global model. This enables collaborative learning in an IIoT system without requiring sensitive data transfers from the PIUs. FL therefore ensures privacy and security, a key concern in IIoT environments, making it an essential technology for advancing IIoT operations \cite{AKmobilityaware}. 

Given the benefits of FL and DNN model compression, this article proposes adopting these strategies to address the aforementioned challenges. Compressed DNNs also further enable the adoption of FL in distributed low-cost PIUs for IIoT applications. We present a case study that demonstrates the efficacy of these methods in an IIoT environment where optimizing PIUs within the FL framework enhances system performance and security. With reduced DNN model size via compression, and privacy achieved via FL, a more robust and effective IIoT system can be achieved.

The rest of the paper is organized as follows: The next section will introduce a high-level discussion of FL. This is followed by a discussion of model compression, specifically addressing pruning and its common variants. We then present a case study that proposes the usage of FL and DNN model compression to acquire compact yet reliable DNN models for PIUs in IIoT. Next we present key future research directions to achieve even more robust and efficient FL industrial IoT systems. Finally, the conclusion of this article is provided.

\section{Federated Learning at the Edge: OTA Aggregation Insights}
The following section introduces the system by discussing federated edge learning and OTA-aggregation.
\subsubsection{Federated Edge Learning}
FL a technique that intertwines wireless communication and ML is central to advancing the industrial IoT era. It involves training a DNN model, known as the global model, which is broadcast by a parameter server (PS) to a multitude of decentralized edge devices such as IoT sensors or PIUs  \cite{caocommunicationefficient}. For each PIU, the received global model now becomes a local model, independently trained with their local dataset. After training, only the gradient vectors (GVs) are sent back to the PS. The broadcasting, local training, and aggregation process continues until the global model reaches a desired level of accuracy, making FL an ideal solution for collaborative yet privacy-preserving computational intelligence techniques in a distributed learning environment.

In an FL network in IIoT, as shown in Figure \ref{Figure main: Federated Learning via Over-the-air Aggregation}, multiple PIUs collaborate to train a DNN model, referred to as the global model. Initially, the PS broadcasts the global model to all PIUs, where each receives an identical version. Subsequently, every PIU trains this model with their respective local datasets. This personalizes the global model into a distinct local model at each PIU.

As each PIU has different data, it trains the model differently, resulting in considerable heterogeneity in training the model. With the completion of local training, the GVs of each local model are now transmitted OTA back to the PS. The PS thus receives an aggregated version of all GVs due to OTA aggregation.

In ML, a crucial concept is the `cost function', a mathematical formulation used to measure the error or difference between the predicted outcomes of a model and the actual results. The FL process aims to find the best set of GVs that minimize this error, hence acquiring a global model that performs optimally in terms of accuracy. As this global model is trained with GVs reflecting the unique data and experiences from each PIU, it becomes more robust, enabling it to adapt effectively to the dynamic environment of industrial setups. 

The process of transmitting and receiving local and global GVs continues until the cost function reaches the lowest possible value. Ideally, this set of GVs shall be achieved in as few FL epochs as possible. Reducing the cost function involves making use of optimization algorithms, one of which is stochastic gradient descent (SGD) \cite{li2014efficient}. Overall, this process highlights the collaborative yet decentralized nature of training a DNN model, where multiple PIUs contribute by maintaining their local data privacy.
\begin{figure}[t]
\renewcommand{\figurename}{\hspace{0cm}Figure}
\centering
\begin{tikzpicture}
\node[draw, inner sep=3pt] {
    \includegraphics[width=\linewidth]{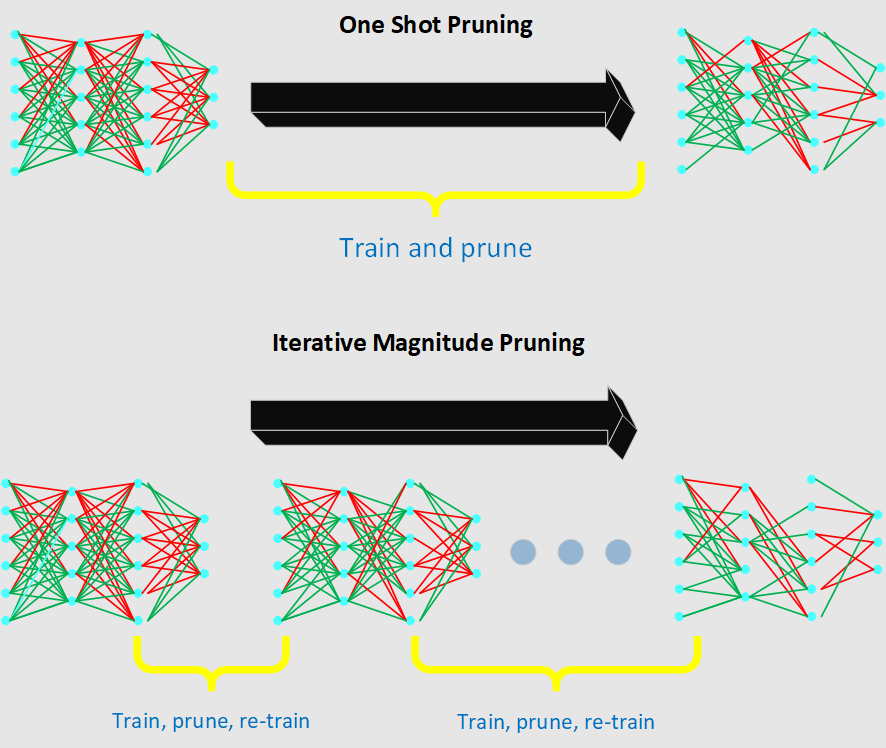}
};
\end{tikzpicture}
\caption{Contrast between OSP and IMP of a DNN. The top sequence shows a network before and after OSP where a significant one-time reduction in connections is obtained after completion of model training. The bottom sequence illustrates the process of IMP, where the model is gradually pruned to the desired ratio, resulting in a selectively pruned network across multiple iterations.}
\label{OSP vs IMP}
\end{figure}
\subsubsection{Over-the-Air (OTA) Aggregation and Wireless Medium} OTA aggregation is an efficient technique for federated learning proposed in wireless networks \cite{zhengOTA}. 
By leveraging the superposition property of wireless channels, OTA aggregation allows the PS to receive a composite signal, combining GVs from multiple PIUs. This process effectively consolidates data transmission, significantly reducing the need for bandwidth and processing and thereby alleviating the computational load at the PS. Particularly useful in settings like manufacturing plants, where PIUs collect vital data such as temperature, this method underscores the practical advantages in industrial environments by streamlining data aggregation and improving the efficiency of distributed learning systems.

In FL, OTA aggregation facilitates the reception of a unified GV set from all participating PIUs without requiring individual transmission resources for each PIU. In the context of the IIoT, where most transmissions involve OTA communication, these data exchanges become particularly susceptible to the complexities of wireless channels. Moreover, the additive white Gaussian noise (AWGN) introduced by PIUs can further degrade DNN model's performance \cite{deepcompression}. 

\begin{figure*}[t!]
\renewcommand{\figurename}{\hspace{0cm}Figure}
	\begin{minipage}{0.50\textwidth}
		\centering
		\hspace*{-0.95cm} 
		\includegraphics[width = 0.9\textwidth]{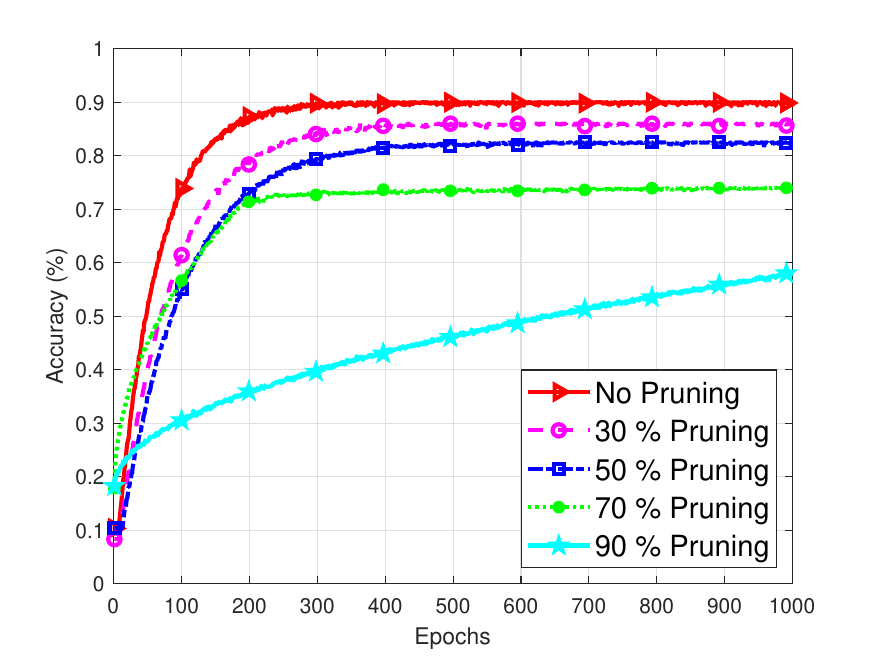}
		\captionsetup{width=.9\linewidth}
		\caption{OSP with full participation}\label{Figure 1: One-shot pruning ResNet18 full participation}    
        \end{minipage}
	\hfill
 	\begin{minipage}{0.50\textwidth}
		\centering
		\hspace*{-0.95cm} 
		\includegraphics[width = 0.9\textwidth]{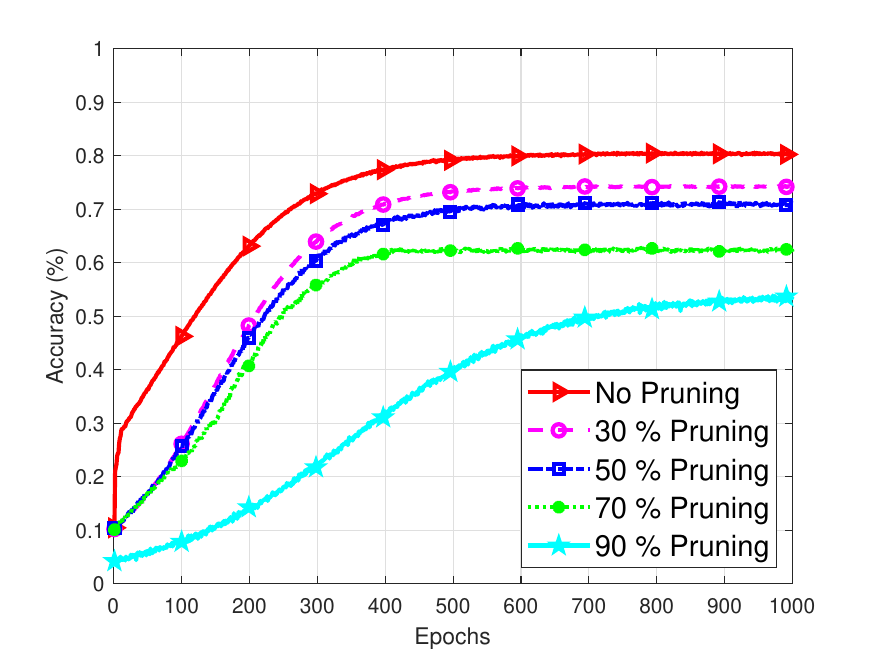}
		\captionsetup{width=.9\linewidth}
		\caption{OSP with partial participation}\label{Figure 2: One-Shot pruning ResNet18 partial participation}    
        \end{minipage}

  \renewcommand{\figurename}{\hspace{0cm}Figure}
 \end{figure*}
 \begin{figure*}[t!]
 \renewcommand{\figurename}{\hspace{0cm}Figure}
         \begin{minipage}{0.50\textwidth}
		\centering
		\hspace*{-0.99cm} 
		\includegraphics[width=0.9\textwidth]{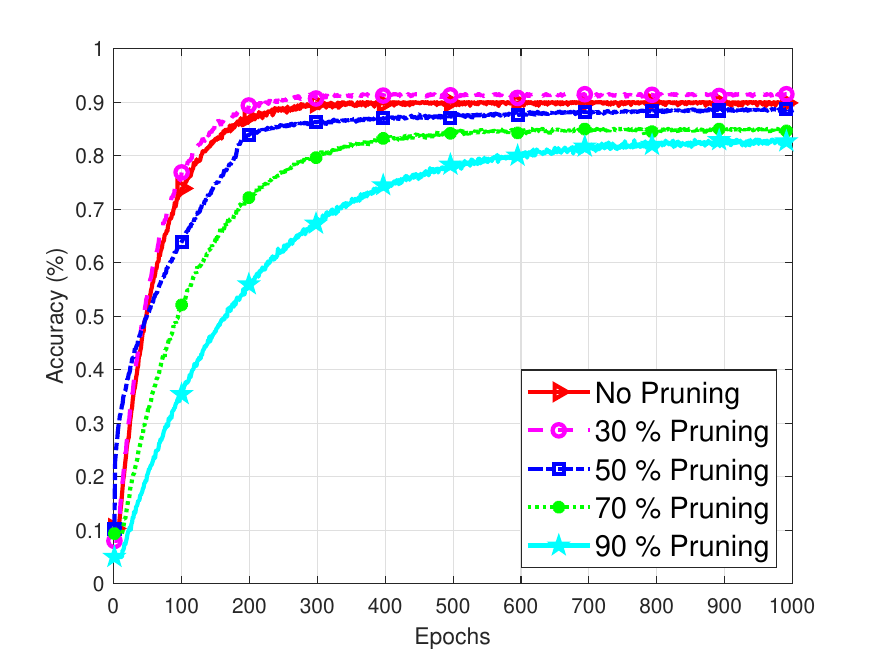}
		\captionsetup{width=.9\linewidth}
		\caption{IMP with full participation}\label{Figure 3: IMP pruning ResNet18 full participation}
	\end{minipage}
	\hfill
 	\begin{minipage}{0.50\textwidth}
		\centering
		\hspace*{-0.99cm} 
		\includegraphics[width=0.9\textwidth]{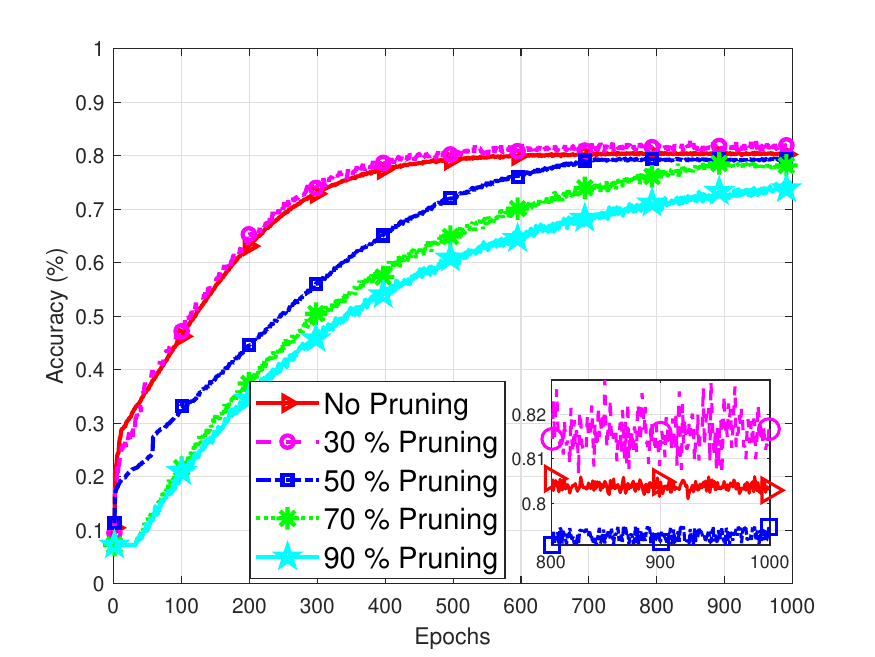}
		\captionsetup{width=.9\linewidth}
		\caption{IMP with partial participation}\label{Figure 4: IMP pruning ResNet18 partial participation}
	\end{minipage}
 \renewcommand{\figurename}{\hspace{0cm}Figure}
\end{figure*}
\section{Deep Neural Network Model Compression: Pruning for Efficiency}
DNN model compression plays a crucial role in enhancing computational efficiency and reducing storage demands, particularly vital for deploying sophisticated ML algorithms on resource-limited PIUs within the IIoT landscape. The primary goal is to streamline DNNs by eliminating redundant parameters, thereby simplifying models without significantly compromising accuracy. Pruning and quantization stand out among compression techniques for their effectiveness in downsizing models while maintaining performance, aligning with the needs of PIUs that operate with constrained memory, bandwidth, and energy \cite{li2023model}.

\subsection{Compression through Pruning}
Pruning cuts down the number of parameters and operations required for making predictions, leading to faster and more energy-efficient inference \cite{ondevice2023}. Having significantly smaller models also enables edge-training in the PIUs which is a fundamental requirement for federated learning. Additionally, pruning helps in identifying and eliminating redundancies within the DNN, potentially enhancing model robustness and reducing the likelihood of overfitting. This makes pruning particularly valuable for IIoT applications where efficiency, reliability, and performance are paramount.

The process of pruning can be broadly categorized into two types: structured and unstructured. Unstructured pruning involves removing individual weights across the DNN based on certain criteria, such as the smallest magnitudes, which leads to a DNN with sparser weights. On the other hand, structured pruning removes entire neurons or channels, leading to reduced dimensions of layers and a more hardware-friendly model structure.

A typical pruning process involves three main steps: training, pruning, and fine-tuning. Initially, the DNN is trained to learn the patterns necessary for performing a specific task. After training, the DNN is pruned by removing weights that contribute the least to the DNN's output based on a predefined metric. Finally, the pruned DNN undergoes fine-tuning, where it is retrained to adjust the remaining weights and recover any lost performance due to pruning. Training and pruning can also happen iteratively as discussed below.

Hence implementing DNN model compression through pruning significantly reduces the size of the DNNs making it an invaluable technique in IIoT where PIUs have limited resources but the cost incurred may be the reduced performance of DNN. Two of the prominent pruning techniques are briefly described below.

\subsubsection{One-shot Pruning (OSP)} OSP involves aggressive pruning of network connections after an initial training period \cite{han2015learning}. The fundamental premise of OSP is based on identifying and removing redundant or less significant connections within a DNN, which is achieved after a comprehensive training process as shown in Figure \ref{OSP vs IMP}. The core idea is that a substantial number of the weights of the DNN model are not critical for maintaining its predictive performance and hence are redundant to the model. By pruning these superfluous connections, OSP significantly reduces the model's size and complexity, leading to lower memory requirements and faster inference time.

Due to its simple one-shot approach, OSP is prone to introducing loss in the model performance at a high pruning percentage. This is because of the pruning of connections that can potentially be important for the DNN. Furthermore, the static nature of OSP implies that it does not adapt or change in response to evolving data or environments, potentially limiting its efficacy in dynamic or complex scenarios.
 
\subsubsection{Iterative Magnitude Pruning (IMP)} Unlike OSP, which prunes a network in a single step via an aggressive approach, IMP  adopts a cyclic process of pruning and retraining to reduce the complexity of a DNN by repeatedly pruning a small percentage of the network's least significant weights \cite{FrankleC19}. This technique involves training a DNN to its full capacity, followed by the incremental removal of the least important weights based on their magnitudes. After each pruning step, the network is retrained to recover performance lost due to the pruning. As shown in Figure \ref{OSP vs IMP}, this cycle of pruning and retraining is iterated several times, allowing the network to adapt progressively and retain its essential connections \cite{pmlr-v119-tan20a}.

The criteria for pruning in each iteration is primarily based on the magnitude of the weights, similar to OSP. The effectiveness of IMP lies in its ability to incrementally discover an optimal subset of connections crucial for maintaining the network's performance. The iterative nature of IMP enables a more nuanced and controlled pruning process, ultimately leading to a more efficient and compact DNN model that can maintain or even surpass the original network performance. 

\section{Case Study: Iterative Magnitude Pruning in OTA-FL for IIoT}
To test the efficacy of IMP in OTA-FL for an IIoT network, this case study utilizes the ResNet18 DNN model which is trained on the CIFAR-10 dataset at the  PIUs. This dataset has a total of $50000$ RGB training and $10000$ test images divided into 10 classes. The dataset is distributed to a total of \textit{E} = $100$ PIUs in an independent and identically distributed (i.i.d.) manner. The performance of the model is tested for both maximum, i.e., full participation (FP) of $100$ PIUs and partial participation (PP) of $50$ PIUs. Both IMP and OSP are performed at each PIU. Further, communication between PIUs and PS stops when the time \textit{t} reaches $1000$ epochs, the upper limit chosen for this setup. The batch size utilized for DNNs in this work is $64$, while the learning rate is $0.001$. The learning curves are provided for a signal-to-noise ratio (SNR) of $10$ dB.
\subsection{Accuracy Results with OSP}
\subsubsection{Full participation}
Figure \ref{Figure 1: One-shot pruning ResNet18 full participation} displays the result for full participation of PIUs, where the unpruned model reaches an accuracy of 90\% while 30P (30\% pruned) and 50P (50\% pruned) manage to attain 86\% and 81\% accuracy, respectively. A significant drop in the accuracy is noted for 70P (70\% pruned) as the model reaches an accuracy of roughly 74\%. Again, among the pruned models, 90P (90\% pruned) roughly reaches 59\% accuracy at the final epoch, which is a significant decrease in the model performance in terms of accuracy. In terms of size reduction, we note a significant reduction as the pruning ratio increases. From unpruned model size of 44.65 Megabytes (MBs), the 30P acquires a size of 32.73 MBs, 50P has a size of 22.39 MBs, while the 70P and 90P manage to acquire sizes of 13.28 and 4.78 MBs, respectively.

\subsubsection{Partial Participation} When the participation from PIUs is reduced to half, i.e., \textit{E} = $50$, the result for accuracy performance for ResNet18 further reduces as compared to the full participation as evident from Figure \ref{Figure 2: One-Shot pruning ResNet18 partial participation}. The unpruned model rests with 80 \% accuracy, while 30P and 50P maintain little difference in accuracies, reaching roughly 73 and 70\% accuracy at the final epoch, respectively. While fairly stable convergence can be noted for 70P, the 50P, with notably slow convergence, reaches an accuracy of 52\%.

\subsection{Accuracy Results with Iterative Magnitude Pruning}
\subsubsection{Full Participation} Figure \ref{Figure 3: IMP pruning ResNet18 full participation} displays remarkable improvement in accuracy for all the pruned versions of ResNet18 DNN when IMP is employed. It can be observed that 30P manages to even surpass the accuracy of the unpruned model. 50P also manages to perform almost at par with the unpruned model, while 70P moves from $74$\% with OSP to almost $85$\%. Similarly, 90P also shows a gain of roughly $23$\% versus basic OSP.

\subsubsection{Partial Participation} A similar improvement in accuracy for all the pruned models can be observed for ResNet18 in Figure \ref{Figure 4: IMP pruning ResNet18 partial participation}, where 30P surpasses the accuracy of the unpruned model. This notes an accuracy improvement of roughly $8$\% from the previous strategy. 50P, though with slower convergence, manages to reach $78$\% accuracy as well. More stable convergence curves are acquired for 70P and 90P which manage to gain an improvement of roughly $16$\% and $20$\%, respectively.

\section{Future Research Directions} 
 \subsubsection{Explainable AI for Effective Pruning} Incorporating explainable AI (XAI) into the compression process in OTA-FL for PIUs can make the pruning results more transparent and understandable \cite{explainableai2022}. This would be particularly interesting for PIUs that gather distinct information and hence have non-i.i.d. data to train the global model. XAI can help identify the most relevant features of the dataset as well as the models for specific tasks, thereby making the compression process more robust. This leads to the creation of efficient and simple models for OTA transmission without sacrificing performance. XAI's role is to clarify the rationale behind compression decisions, enhancing the trust and reliability of the resulting compressed models.

\subsubsection{Adaptive Compression} OTA-FL and compression techniques can be tested further by employing an adaptive data handling strategy that considers the unique characteristics and heterogeneity of each PIUs dataset in the network and adjusts the compression rate accordingly. This could involve deeper pruning or other compression techniques, e.g., quantization, to ensure that areas of the model sensitive to specific data distributions undergo minimal compression to maintain the accuracy of the global model. This can be done by developing a feedback loop that assesses the impact of compression on the model's effectiveness in real time, allowing on-the-fly adjustments to compression techniques. Further, dynamically updating the probability of a PIU being selected for model training based on a metric called cumulative model strength, which measures the statistical heterogeneity and previous training performance of a PIU, will help improve the overall network performance in IIoT.

\subsubsection{Retraining at PS} Another prospective future work would be the incorporation of model retraining at the PS. In this strategy, the PS may have access to some samples of data that is used for local training. The PIUs after training the model transmit the GVs to PS,  where the PS updates the model with received GVs and retrains it with the data it has access to. This retraining of the global model can make the model more robust, hence enhancing its performance, particularly for PIUs that operate in non-i.i.d., setups. The requirement, however, is that the PS can access some generic representative data. In this approach, PIUs can also locally compress and fine-tune their models to reduce size and complexity before transmitting them for aggregation. 

\subsubsection{Multi Agent Reinforcement Learning for PIUs Selection} The work can further be extended with multi-agent reinforcement learning (MARL) for efficient PIUs selection. By independently learning optimal compression strategies, MARL-based frameworks can optimize PIUs selection, enhancing model accuracy while reducing processing latency and communication costs in IIoT \cite{deepreinforcAK}. This can produce robust communication protocols that involve efficient model updates based on the noisy channel condition and the accuracy affected by deeper compression. Each edge device, functioning as an agent, learns to adapt its compression strategy based on individual data and industrial conditions. This would optimize local model updates and result in better performance of the OTA-FL network in IIoT. One particular challenge will include ensuring efficient learning across agents handling the variability of wireless channels in OTA scenarios, designing a practical reward function, and ensuring efficient communication among agents.

\subsubsection{Investigating Performance with More Complex Tasks} 
As industrial setups are becoming more diverse with the presence of various formats of data, sophisticated algorithms, and increasingly complex DNN models are being proposed to handle them effectively. Among these techniques, video monitoring has emerged as a crucial tool for quality control and surveillance within industrial frameworks. However, employing complex DNN models like 3D convolutional networks (3D-CNNs) for video processing introduces significant challenges due to the complexity of video data, resulting in 3D-CNNs with substantially large number of parameters. Future research could focus on compression techniques that effectively reduce the volume of these complex DNN models, before transmitting their parameters OTA. This would significantly lower the demand for network resources while retaining the essential quality and utility of the data for critical functions such as activity recognition. In-depth exploration and enhancement of these data compression strategies could lead to substantial improvements in bandwidth efficiency and processing speed, particularly in OTA-FL scenarios and broader industrial IoT contexts.

\section{Conclusion}
Industrial IoT enables a plethora of applications across sectors with improved safety and security of information exchange among PIUs, and other operational efficiencies. In this article, we provided a detailed overview to understanding the role of model pruning in the context of OTA federated learning without sacrificing model performance. Following this, we delved into a case study that showcases the practical benefits and potential opportunities of applying IMP within an OTA-FL network.  To reduce the DNN model in a bid to save bandwidth and energy of PIUs, simple one-shot pruning can be utilized. However, one-shot pruning results in a considerable performance loss, particularly in the OTA-FL context where channel noise is problematic. To overcome this loss in performance we demonstrate how iterative magnitude pruning can be used. The results show considerable gains from IMP to regain the accuracy lost with OSP, particularly with full participation of the PIUs for model training. Finally, we outlined future research opportunities and directions that can further improve the usefulness of model pruning and federated learning for IIoT networks.

\bibliographystyle{IEEEtran}
\bibliography{IMP.bib}
\end{document}